\documentclass[a4paper,conference]{IEEEtran}
\usepackage{cite}
\usepackage{tabularx,booktabs}
\usepackage{multirow}

\usepackage[colorlinks=true,linkcolor=blue,citecolor=blue]{hyperref}%
\usepackage[pdftex]{graphicx}

\usepackage{amsmath}

\usepackage{algorithm}

\usepackage{algorithmic}
\usepackage{hyperref}
\hypersetup{
    colorlinks=true,
    linkcolor=blue,
    filecolor=magenta,      
    urlcolor=blue,
}

\usepackage[caption=false,font=normalsize,labelfont=sf,textfont=sf]{subfig}

\usepackage{dblfloatfix}
\newtheorem{theorem}{Theorem}

\newtheorem{definition}{Definition}[section]
\newtheorem{Proposition}[theorem]{Proposition}

\usepackage{url}
\renewcommand\IEEEkeywordsname{Keywords}

\begin{document}
\title{Aggregating Dependent Gaussian Experts in Local Approximation}
\author{\IEEEauthorblockN{Hamed Jalali}
\IEEEauthorblockA{Department of Computer Science\\
University of Tuebingen, Germany\\
Email: hamed.jalali@wsii.uni-tuebingen.de}
\and
\IEEEauthorblockN{Gjergji Kasneci}
\IEEEauthorblockA{Department of Computer Science\\
University of Tuebingen, Germany\\
Email: gjergji.kasneci@uni-tuebingen.de}
%\and
%\IEEEauthorblockN{James Kirk\\ and Montgomery Scott}
%\IEEEauthorblockA{Starfleet Academy\\
%San Francisco, California 96678--2391\\
%Fax: (888) 555--1212}
}

\maketitle

\begin{abstract}
Distributed Gaussian processes (DGPs) are prominent local approximation methods to scale Gaussian processes (GPs) to large datasets. Instead of a global estimation, they train local experts by dividing the training set into subsets, thus reducing the time complexity. This strategy is based on the conditional independence assumption, which basically means that there is a perfect diversity between the local experts. In practice, however, this assumption is often violated, and the aggregation of experts leads to sub-optimal and inconsistent solutions. In this paper, we propose a novel approach for aggregating the Gaussian experts by detecting strong violations of conditional independence. The dependency between experts is determined by using a Gaussian graphical model, which yields the precision matrix. The precision matrix encodes conditional dependencies between experts and is used to detect strongly dependent experts and construct an improved aggregation. Using both synthetic and real datasets, our experimental evaluations illustrate that our new method outperforms other state-of-the-art (SOTA) DGP approaches while being substantially more time-efficient than SOTA approaches, which build on independent experts.
\end{abstract}
\vspace{2mm}

\iffalse
\IEEEpeerreviewmaketitle
\fi

\section{Introduction}
Gaussian processes (GPs) \cite{Rasmussen} are flexible, interpretable, and powerful non-parametric statistical methods which provide accurate prediction with a low amount of uncertainty. They apply Bayes' theorem for inference, which allows them to estimate complex linear and non-linear structures without the need for restrictive assumptions of the model. They have been extensively used in practical cases, e.g. optimization \cite{Shahriari}, data visualization, and manifold learning \cite{Lawrence}, reinforcement learning \cite{Deisenroth2013}, multitask learning \cite{Alvarez}, online streaming models \cite{Huber, Le}, and time series analysis \cite{Petelin, Tobar}. The main bottleneck of using standard GPs is that they poorly scale with the size of the dataset. For a dataset of size \textit{N},  the training complexity is $\mathcal{O}(N^3)$ because the inversion and determinant of the $N\times N$ kernel matrix are needed. The prediction over a test set and also storing the results suffers from an additional complexity of $\mathcal{O}(N log N)$. This issue currently restricts GPs to relatively small training datasets, the size of which is typically in the order of $\mathcal{O}(10^4)$.

To deal with large datasets two different strategies are used. The first strategy is based on sampling a small subset of the full dataset. The methods that follow this strategy try to train a GP on the smaller subset and then generalize the results. A simple method in this case, is subset-of-data (SoD) \cite{Chalupka} which only uses a subset of size \textit{m} from the original dataset; its training complexity is $\mathcal{O}(m^3)$ where $m<<N$. Since this approach ignores the remaining data, it has limited performance. 

Another method which is called sparse kernel or compactly supported kernel \cite{Gneiting,Melkumyan}, ignores the observations that are not correlated or show a covariance that is smaller than a threshold. In radial-based kernels, if the distance between two different entries is larger than a determined value, their covariance are set to zero. Although the training complexity of this method is $\mathcal{O}(\alpha N^3)$, for certain interesting cases, it does not guarantee that the new modified kernel is positive semi-definite.

The most popular method in this area is the sparse approximation approach, which employs a subset of the data (called inducing points) and \textit{Nyström} approximation to estimate the prior and posterior distributions \cite{Titsias, Hensman}. For \textit{m} inducing points, the training complexity is $\mathcal{O}(Nm^2)$. Although the authors provide a full probabilistic model using the Bayesian framework, it is not conceivable to apply the method to large and high dimensional datasets because its capability is restricted by the number of the inducing points \cite{Bui, Moore}.

The second strategy is to divide the full dataset into partitions, train the local GPs in each partition \cite{Snelson, Urtasun}, and then aggregate the local approximations \cite{Cao, Deisenroth2015, Liu}. Unlike sparse approximations, this local approach can model quick-varying systems and non-stationary data features. Since in this family, the training procedure is run in different subsets, the final prediction may be affected by the regions with the poor predictive performance or by discontinuous predictions in overlapping sub-regions.

The most popular local approximation methods are the mixture of experts (MoE) and the product of experts (PoE). The MoE works as a Gaussian mixture model (GMM). It combines the local experts with their hyper-parameters and improves the overall predictive power \cite{Tresp2001, Masoudnia}. The main drawback of this method is that a joint training is needed to learn the mixing probabilities and the individual experts. This joint training positively affects the predictive power and helps control the experts with poor performance, but - on the negative side - it increases the complexity \cite{Cao}.   

The prominent product of experts (PoE) \cite{Hinton} and Bayesian committee machine (BCM) \cite{Tresp} provide a new framework for GPs. Independent experts are GPs that are learned separately. Both methods suffer from the discontinuity issue and the weak experts' problem \cite{Camps-Valls, Liu}. The generalized product of experts (GPoE) \cite{Cao} and robust Bayesian committee machine (RBCM) \cite{Deisenroth2015} propose different aggregation criteria, which are robust to weak experts' predictions.

To cope with the consistency problem in the predictions, \cite{Rulli'ere} suggested the nested pointwise aggregation of experts (NPAE), which provides consistent predictions but increases the time complexity. The authors of \cite{Liu} proposed the generalized robust Bayesian committee machine (GRBCM) by considering one expert as a base expert, i.e., a global expert that is modified the RBCM to provide consistent predictions. The authors in \cite{Liu} showed that this modified RBCM is capable of providing consistent predictions, especially for the disjoint data partitioning regime. 

The idea behind the  BCM and PoE families of methods is the conditional independence (CI) assumption between the local experts. These divide-and-conquer approaches can speed up the computation and provide a distributed learning framework. However, since the CI assumption is violated in practice, they return poor results in cases with dependent experts.

The key contribution of our work lies in considering the dependency between Gaussian experts and improve the prediction quality in an efficient way. To this end, we first develop an approach to detect the conditional correlation between Gaussian experts, and then we modify the aggregation using this knowledge. In the first step, a continuous form of a Markov random field is used to infer dependencies and then the expert set is divided into clusters of dependent experts. In the second step, we adopt GRBCM for this new scenario and present a new aggregation method that is accurate and efficient and leads to better predictive performance than other SOTA approaches, which use the CI assumption. 

The structure of the paper is as follows. Section II introduces the GP regression problem and SOTA DGP approaches. In Section III the proposed model and the inference process are presented. Section IV shows the experimental results, and we conclude in Section V.

\section{Problem Set-up}

\subsection{Background}

Let us consider the regression problem $y=f(x)+\epsilon$, where $x\in R^D$ and $\epsilon \sim \mathcal{N}(0,\sigma^2)$, and the Gaussian likelihood is $p(y|f)=\mathcal{N}(f, \sigma^2 I)$. The objective is to learn the latent function \textit{f} from a training set $\mathcal{D}=\{X,y\}_{i=1}^n$. The Gaussian process regression is a collection of function variables any finite subset of which has a joint Gaussian distribution. The GP then describes a prior distribution over the latent functions as $f \sim GP\left(m(x),k(x,x^{'}) \right)$, where $m(x)$ is a mean function and $k(x,x^{'})$ is the covariate function (kernel) with hyperparameter $\psi$. The prior mean is often  assumed as zero, and the kernel is the 
well-known squared exponential (SE) covariance function equipped with automatic relevance determination (ARD), 
\[k(x,x^{'})=\sigma_f^2 \; \exp\left( -\frac{1}{2} \sum_{i=1}^D \frac{(x_i-x_i^{'})^2}{\mathcal{L}_i} \right)\]
where $\sigma_f^2$ is a signal variance, and $\mathcal{L}_i$ is an input length-scale along the \textit{i}th dimension, and $\psi=\{\sigma_f^2,\mathcal{L}_1,\ldots,\mathcal{L}_D\}$. To train the GP, the hyper-parameters $\theta=\{\sigma^2, \psi\}$ should be determined such that they maximise the log-marginal likelihood \cite{Rasmussen}
\begin{equation} \label{eq:gp_likelihood}
\log\, p(y|X)=-\frac{1}{2}y^T\mathcal{C}^{-1}y - \frac{1}{2}\log|\mathcal{C}|- \frac{n}{2}\log(2\pi)
\end{equation}
where $\mathcal{C}=K+\sigma^2I$. For a test set $x^*$ of size $n_t$, the predictive distribution is also a Gaussian distribution $p(y^*|D,x^*)\sim \mathcal{N}(\mu^*,\Sigma^*)$, with mean and covariance respectively given by
\begin{equation} \label{eq:gp_mean}
\mu^*=k_*^T(K+\sigma^2I)^{-1}y,
\end{equation}
\begin{equation} \label{eq:gp_var}
\Sigma^*=k_{**} - k_*^T(K+\sigma^2I)^{-1}k_*,
\end{equation}
where $K=k(X,X)$, $k_*=k(X,x^*)$, and $k_{**}=k(x^*,x^*)$.  

According to \eqref{eq:gp_likelihood}, the training step scales as $\mathcal{O}(n^3)$ because it is affected by the inversion and determinant of $\mathcal{C}$, which is an $n \times n$ matrix. Therefore, for large datasets, training is a time-consuming task and imposes a limitation on the scalability of the GP.

\subsection{Distributed Gaussian Process}

To scale the GP to large datasets, the cost of the standard GP is reduced by distributing the training process. It involves dividing the full training dataset $\mathcal{D}$ into $M$ partitions $\mathcal{D}_1,\ldots,\mathcal{D}_M$, (called experts) and training the standard GP on these partitions. The predictive distribution of the i'th expert $\mathcal{M}_i$ is $p_i(y^*|\mathcal{D}_i,x^*)\sim \mathcal{N}(\mu_i^*,\Sigma_i^*)$, where its mean and variance are calculated by using \eqref{eq:gp_mean} and \eqref{eq:gp_var} respectively
\begin{equation} \label{eq:experts_mean}
\mu_i^*=k_{i*}^T(K_i+\sigma^2I)^{-1}y_i,
\end{equation}
\begin{equation} \label{eq:experts_var}
\Sigma_i^*=k_{**} - k_{i*}^T(K_i+\sigma^2I)^{-1}k_{i*}.
\end{equation}

Aggregating these experts is based on the assumption that they are independent. The most prominent aggregation methods are PoE \cite{Hinton} and BCM \cite{Tresp}. GPoE \cite{Cao} and RBCM \cite{Deisenroth2015} are new modified versions of PoE and BCM, which approach the discontinuity problem and overconfident predictions.

The term distributed Gaussian process was proposed by \cite{Deisenroth2015} to include PoE, BCM, and their derivatives, which are all based on the fact that the computations of the standard GP is distributed amongst individual computing units. Unlike sparse GPs, DGPs make use of the full dataset but divide it into individual partitions.

The predictive distribution of DGP is given as the product of multiple densities (i.e., the experts). If the experts $\{\mathcal{M}\}_{i=1}^M$ are independent, the predictive distribution of DGP for a test input $x^*$ is 
\begin{equation} \label{eq:gpoe}
p(y^*|\mathcal{D},x^*) \propto \prod_{i=1}^M p_i^{\beta_i}(y^*|\mathcal{D}_i,x^*).
\end{equation}
The weights $\beta = \{\beta_1,\ldots,\beta_M\}$ describe the importance and influence of the experts. The typical choice of the weights is the difference in differential entropy between the prior $p(y^*|x^*)$ and the posterior $p(y^*|\mathcal{D},x^*)$ \cite{Cao}. With such weights however, the predictions of GPoE are too conservative and the predictions are not appropriate \cite{Liu}. To address this issue, the simple uniform weights $\beta_i=\frac{1}{M}$ is used \cite{Deisenroth2015}. The predictive distribution of GPoE with normalized weights asymptotically converges to the full Gaussian process distribution but is too conservative \cite{Szabo}.

\subsection{Discussions of the Properties Existing Aggregations}
\paragraph{Consistency}
To deal with the inconsistency issue, the nested pointwise aggregation of experts (NPAE) \cite{Rulli'ere} considers the means of the local predictive distributions as random variables by assuming that $y_i$ has not been observed and therefore allows the dependency between individual experts' predictions. Theoretically, it provides consistent prediction but its aggregation steps need much higher time complexity. For M individual partitions, NAPE needs to calculate the inverse of a $M \times M$ matrix in each test point that leads to a longer running time when a large training set is used or the number of partitions is large.

Another new model is the generalized robust Bayesian committee machine (GRBCM)~\cite{Liu} which introduces a base (global) expert and considers the covariance between the base and other local experts. For a global expert, $M_b$, in a base partition, ${D}_{b}$, the predictive distribution of GRBCM is
\begin{equation} \label{eq:grbcm}
p(y^*|\mathcal{D},x^*)= \frac{\prod_{i=2}^M p_{bi}^{\beta_i}(y^*|\mathcal{D}_{bi},x^*)}{p_b^{\sum_{i=2}^M \beta_i-1}(y^*|\mathcal{D}_b,x^*)} ,
\end{equation}
where $p_b(y^*|\mathcal{D}_b,x^*)$ is the predictive distribution of $M_b$, and $p_{bi}(y^*|\mathcal{D}_{bi},x^*)$ is the predictive distribution of an expert trained on the dataset $\mathcal{D}_{bi}=\{\mathcal{D}_{b},\mathcal{D}_{i}\}$.
It improves the prediction and consistency of the RBCM has time complexity $\mathcal{O}(\alpha nm^2_0) + \mathcal{O}(\beta n^{'}n m_0)$, where $m_0$ in the number of assigned points to each expert, $n^{'}$ is the size of test set, $\alpha=(8M-7)/M$, and $\beta=(4M-3)/M$\cite{Liu}. \\

\paragraph{Conditional independence (CI)}
CI is a crucial assumption for many unsupervised ensemble learning methods. It has been used widely in regression and classification problems \cite{Moreira,Parisi}. The (R)BCM and (G)PoE methods are also based on CI assumption which reduces the computational costs of the training process, see Figure \ref{fig.1.a}. 
 
 However, in practice, this assumption is often violated and their predictions are not accurate enough. Actually, the ensembles based on CI return sub-optimal solutions \cite{Jaffe}. In this regard, few works have considered modelling dependencies between individual predictors. For classification, \cite{Donmez} used pairwise interaction between classifiers and \cite{Platanios} considered the agreement rates between subsets of experts. In another work~\cite{Jaffe}, the authors suggested a model using clusters of binary classifiers in which the classifiers in each cluster are conditionally dependent. The authors of~\cite{Jaffe} defined a specific score function based on covariance between classifiers to detect dependency.  
 
 In local approximation GPs, the only method that considered the dependency between experts is NPAE\cite{Rulli'ere, Bachoc}. It assumes the joint distribution of experts and $y^*$ is a Gaussian distribution and uses the properties of conditional Gaussian distributions to define the meta-learner. Due to the high computational cost which cubically depends on the number of experts at each test point, this method does not provide an efficient solution for large real-world datasets.
 
 In the next section, we propose a new model that uses the dependency between experts and define a modified aggregation method based on GRBCM.
 
\section{Distributed Gaussian Process with Dependent Experts}
\begin{figure}[hbt!]
\centering
\subfloat[]{\includegraphics[width=0.8\columnwidth]{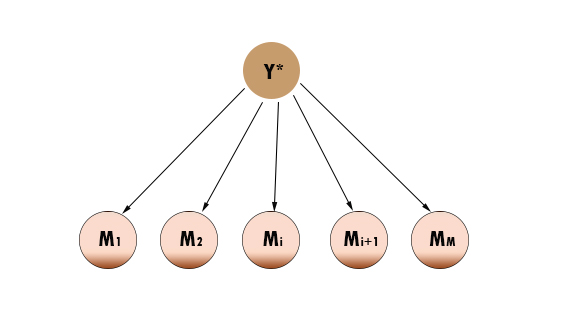}%
\label{fig.1.a}}
\hfil
\subfloat[]{\includegraphics[width=0.8\columnwidth]{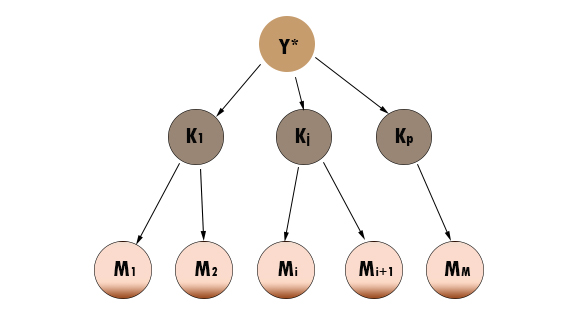}%
\label{fig.1.b}}
\caption{ \textbf{Computational graphs}: (a) DGP model with CI \cite{Deisenroth2015}; (b) DGP with clusters of dependent experts \cite{Jaffe}}
\label{fig.1}
\end{figure}

Assume the Gaussian experts $\mathcal{M}=\{\mathcal{M}_1,\ldots,\mathcal{M}_M\}$ have been trained on different partitions and let $\mu_{\mathcal{M}}^*=[\mu_1^*,\ldots,\mu_M^*]^T$ be a $ n_t \times M$ matrix that contains the local predictions of $M$ experts at $n_t$ test points. Our approach makes use of the experts' predictions, i.e. $\mu_{\mathcal{M}}^*$ in order to detect strong dependencies between experts. This step results clusters of correlated experts, $\mathcal{C}=\{\mathcal{C}_1,\ldots,\mathcal{C}_P\}$, $P \ll M$. By aggregating the experts at each cluster, it leads to a new layer of experts, $\mathcal{K}=\{\mathcal{K}_1,\ldots,\mathcal{K}_P\}$,  which are conditionally independent given $y^*$. Figure \ref{fig.1.b} depicts this model where the experts in cluster $\mathcal{C}_i$  are conditionally independent given $\mathcal{K}_i$, and each $\mathcal{K}_i$ is independent of $\mathcal{K}_j, i\neq j$ given $y^*$. The final prediction is done by using the $\mathcal{K}$ instead of $\mathcal{M}$.\vspace{2mm}
\begin{definition}[\textbf{Assignment function}] \label{def.1}
A function $\mathcal{H}: \mathcal{M} \to \mathcal{C}$ is the assignment function that represents the related cluster for each expert. $\mathcal{H}(\mathcal{M}_i)=\mathcal{C}_j$ means that the $i$'th expert belongs to $j$'th cluster of experts and it has a dependency with the experts in this cluster. Therefore, if $\mathcal{H}(\mathcal{M}_i)=\mathcal{H}(\mathcal{M}_j)$, the $i$'th and $j$'th experts are correlated to each other and belong to the same cluster.
\end{definition}

In the following, we will show how to detect subsets of strongly dependent experts and present a new aggregation method for DGPs. 

\subsection{Dependency Detection with Gaussian Graphical Models}
The key idea in an undirected graphical model (or pairwise Markov random field (MRF)) is to model the set of local estimators as a connected network such that each node represents a Gaussian expert and the edges are the interaction between them. This network model uses a matrix of parameters to encode the graph structure. In other words, it considers the edges as parameters, such that if there is a connection between two nodes, then there are non-zero parameters for the pair.

\paragraph{Gaussian graphical models (GGMs).}
Gaussian graphical models (GGMs) \cite{Rue,Uhler,Drton} are continuous forms of pairwise MRFs, i.e. the nodes are continuous. The basic assumption for GGMs is that the variables in the network follow a multivariate Gaussian distribution. The distribution for GGMs is 
\begin{equation} \label{eq:ggm_variance}
 p(Z|\mu, \Sigma)= \\ \frac{1}{(2\pi)^{Q/2}|\Sigma|^{1/2}} \exp\left\{-\frac{1}{2}(Z-\mu)^T\Sigma^{-1}(Z-\mu) \right\},
\end{equation}
where $Z=\{Z_1,\ldots,Z_Q\}$ are the variables (nodes), Q is the number of variables, and $\mu$ and $\Sigma$ are the mean and covariance, respectively. The distribution in \eqref{eq:ggm_variance} can also be expressed using the precision matrix $\Omega$:   
\begin{equation} \label{eq:ggm_precision}
\begin{split}
p(Z|h, \Omega)=& \frac{|\Omega|^{1/2}}{(2\pi)^{Q/2}} \exp\left\{-\frac{1}{2}(Z-h)^T\Omega(Z-h) \right\} \\ \propto & \exp\left\{-\frac{1}{2}Z^T\Omega Z + h^T Z \right\} ,
\end{split}
\end{equation}
where $\Omega=\Sigma^{-1}$ and $h=\Omega \mu$. The matrix $\Omega$ is also known as the potential or information matrix. 

WLOG, let $\mu= 0$, then the distribution of a GGM shows the potentials defined on each node $i$ as $exp\{-\Omega_{ii}(Z_i)^2\}$ and on each edge $(i,j)$ as $exp\{-\Omega_{ij} Z_i Z_j\}$. 

Unlike correlation networks, Eq.\ \eqref{eq:ggm_variance},  which encode the edge information in the network on the covariance matrix, a GGM is based on the precision matrix, Eq.\ \eqref{eq:ggm_precision}. 
In a correlation network, if $\Sigma_{ij}=0$, then $Z_i$ ard $Z_j$ are assumed to be independent. While in a GGM, if $\Omega_{ij}=0$, then $Z_i$ ard $Z_j$ are conditionally independent given all other variables, i.e. there is no edge between $Z_i$ ard $Z_j$ in the graph.\\

\paragraph{Network Learning.} 
In the network, the locally trained experts are the nodes, and network learning results in the precision matrix $\Omega$; the latter reveals the conditional dependencies between experts. GGMs use the common sparsity assumption, that is, there are only few edges
in the network and thus the parameter matrix is sparse. This assumption usually makes sense in experts' networks because the interaction of one expert is limited to only a few other experts. 

To this end, the Lasso regression \cite{Hastie2015} is used to perform neighborhood selection for the network. The Meinshausen-Bühlmann algorithm \cite{Meinshausen2006} is one of the first algorithms in this area. \cite{Meinshausen2006} and \cite{Wainwright} proved that with some assumptions, Lasso asymptotically recovers correct relevant subsets of edges. \cite{Friedman2008} proposed the efficient graphical Lasso which adopts a maximum likelihood approach subject to an L1 penalty on the coefficients of the precision matrix. The graphical Lasso has been improved in later works \cite{Friedman2010,Friedman2011,Hallac}. 

Let $S$ be the sample covariance. Then, the Gaussian log-likelihood of the precision matrix $\Omega$ is equal to $log |\Omega| - trace(S \Omega)$. The Graphical Lasso (gLasso) maximizes this likelihood subject to an element-wise $L_1$ norm penalty on $\Omega$. Precisely, the objective function is
\begin{equation} \label{eq:glasso}
\hat{\Omega}= \arg\max_{\Omega} \log |\Omega| - trace(S \Omega) - \lambda \left\Vert \Omega \right\Vert_1,
\end{equation}
where the estimated neighborhood is then the non-zero elements of $\hat{\Omega}$. Since $\hat{\Omega}$ contains all information about the dependency between experts, we use it to construct the assignment function $\mathcal{C}$ and the clusters of experts $\mathcal{K}$. 

\subsection{Aggregation}
After determining the dependencies between the experts, we apply the following aggregation method. First, we define clusters of interdependent experts, i.e. that include experts with strong dependency. Then, by using the GRBCM method in $i$th cluster, we generate for each cluster a modified expert $\mathcal{K}_i$. The final prediction is done by aggregating the predictions of these modified experts.

\paragraph{Experts clustering} After detecting the dependencies between experts, we use the precision matrix to find the assignment function. Performing a clustering approach on the precision matrix returns the clusters of experts $\mathcal{K}$; thus, each cluster $\mathcal{K}_i$ contains strongly dependent experts based on the precision matrix. To this end, we apply \emph{spectral clustering (SC)} \cite{Luxburg} which is more robust and works better in practice. Spectral clustering makes use of the relevant eigenvectors of the Laplacian matrix of the similarity matrix (here the precision matrix) alongside a standard clustering method. The Laplacian matrix is $L=D-\Omega$, where $D$ is a diagonal matrix that includes the sum of the values in each row of $\Omega$.  

\begin{figure}[hbt!]
\centering
\subfloat[ $\lambda =0.1$]{\includegraphics[width=0.8\columnwidth]{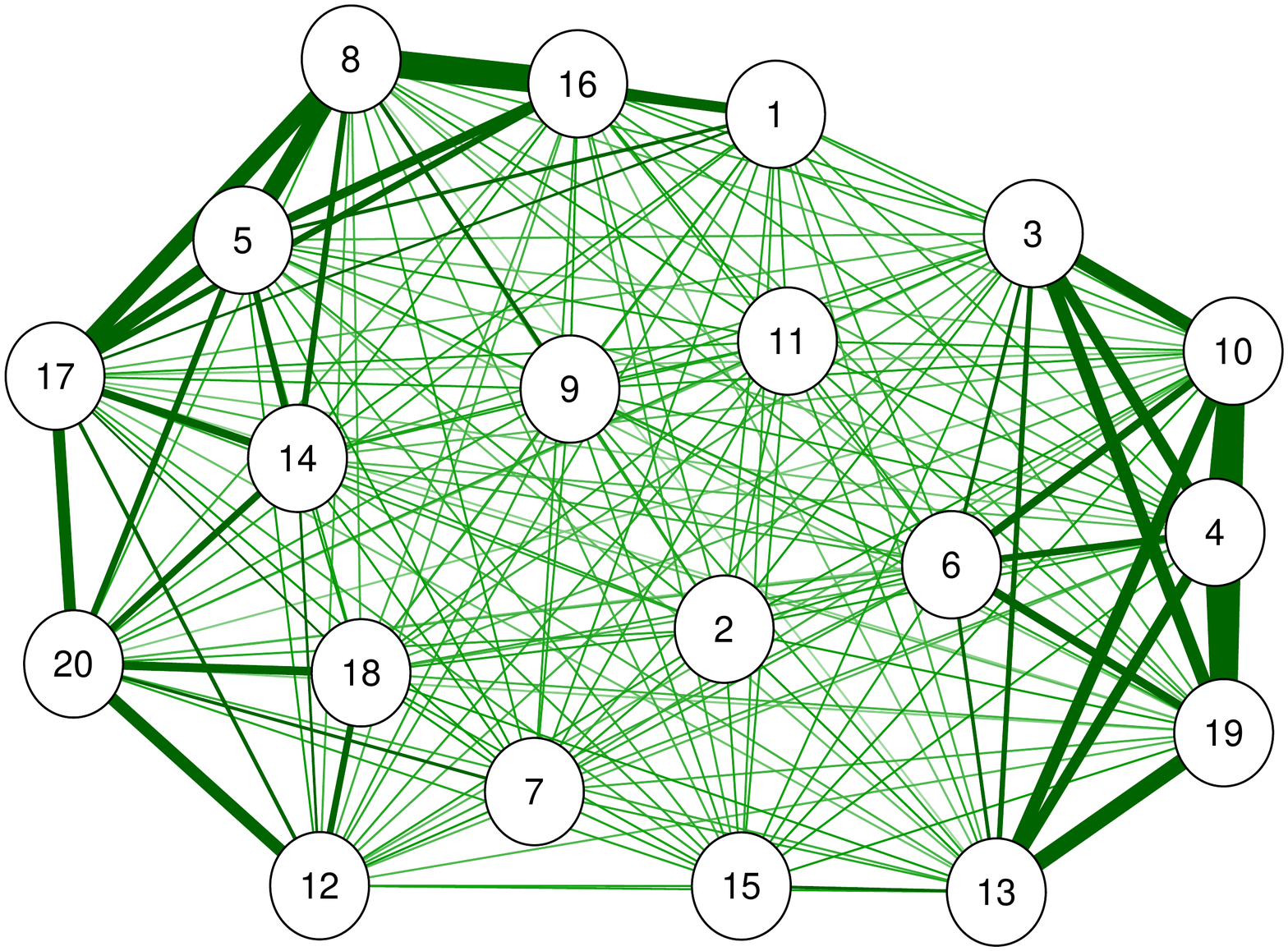}%
\label{fig.2.a}}
\hfil
\subfloat[$\lambda =0.1, P=6$]{\includegraphics[width=0.8\columnwidth]{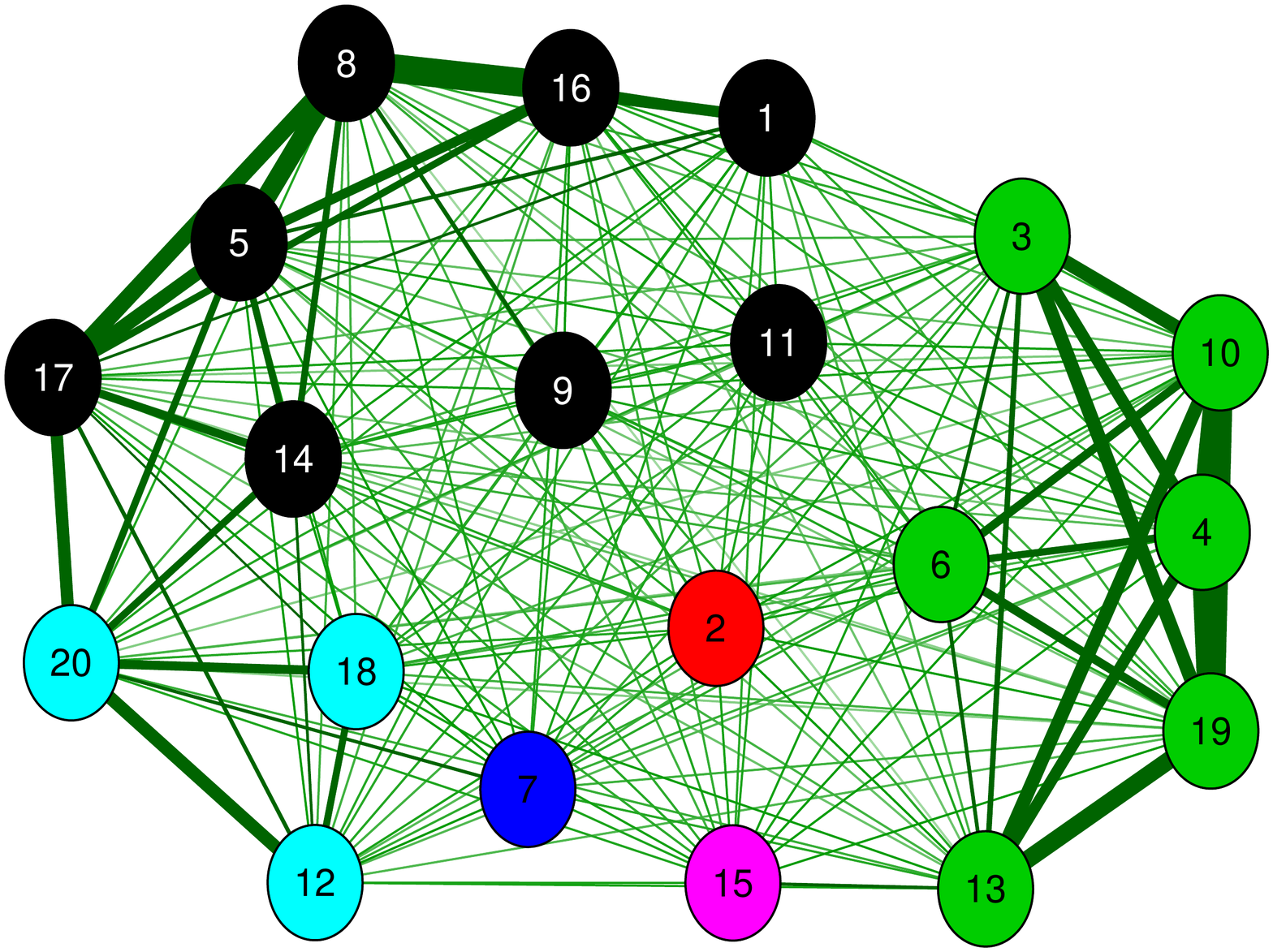}%
\label{fig.2.b}}
\hfil
\subfloat[$\lambda =0.1$]{\includegraphics[width=\columnwidth]{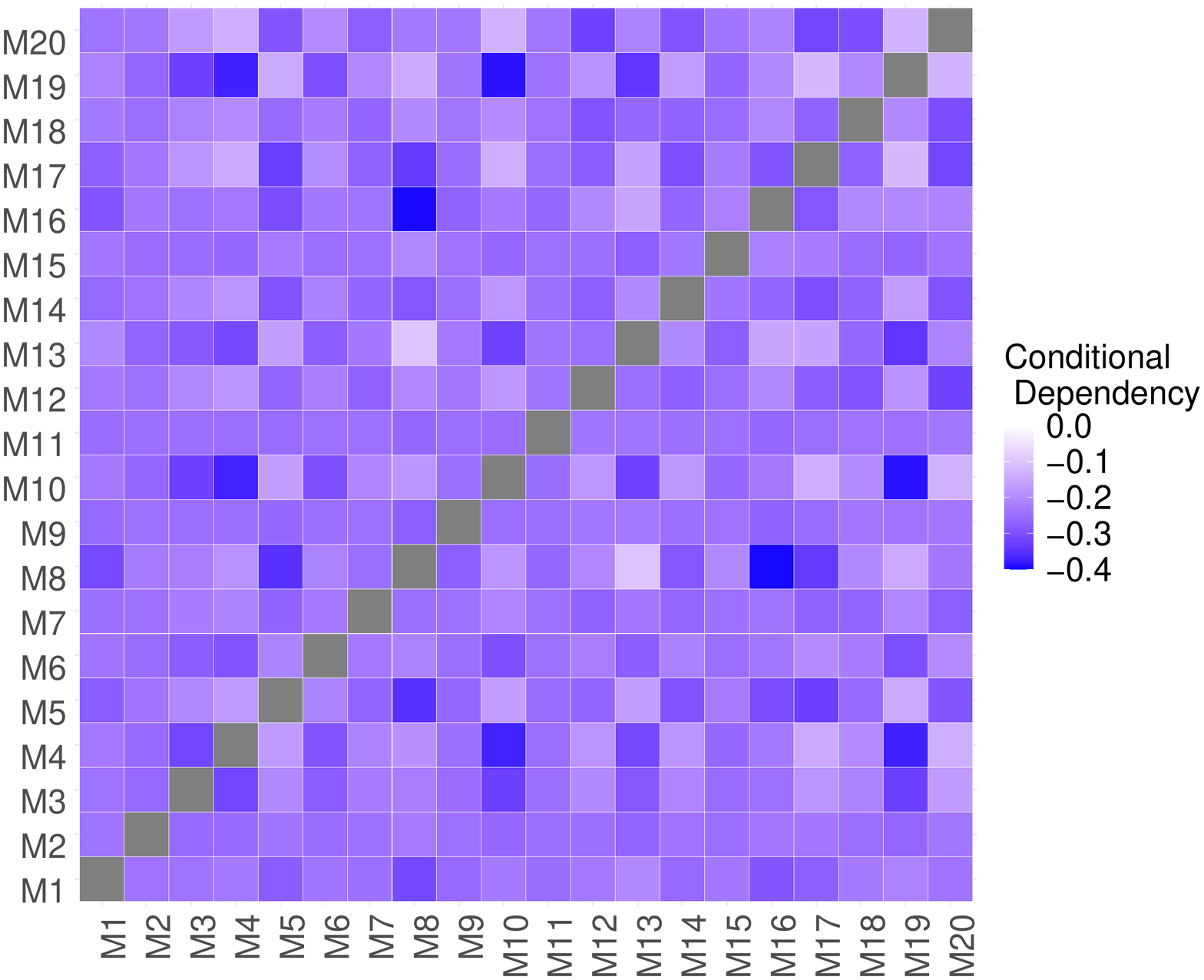}%
\label{fig.2.c}}
\caption{\textbf{Gaussian graphical models}:(a) shows the interaction between experts in a GGM of 20 experts with a penalty term $\lambda =0.1$, (b) reveals the GGM with 6 clusters of experts, and (c) depicts the \emph{heat map} plot of the experts' precision matrix.}
\label{fig.2}
\end{figure}

Figure \ref{fig.2} depicts the GGM of a simulated dataset with $10^5$ training points which have been divided into 20 partitions (experts). This dataset is considered in Section \ref{f_x} in detail. Figure \ref{fig.2.a} represents the sparse graph with a penalty term $\lambda =0.1$ in graphical Lasso with the nodes (experts) and edges (interactions or dependencies). Even with this penalty term, the CI assumption is violated because all experts are connected to each other. Figure \ref{fig.2.b} displays the graph after performing the spectral clustering on the precision matrix. The 6 clusters in the graph contain correlated experts, and clusters that are now represented by only one expert (e.g. the cluster with red color) contain the original experts that are not strongly dependent with the other experts. Figure \ref{fig.2.c} represents the \emph{heat map} plot of the symmetric precision matrix and shows the conditional dependencies between experts. The main diagonal returns the experts internal potential while the other elements are conditional dependencies between experts. In fact this figure shows that the experts are conditionally dependent and the CI assumption is violated.

\paragraph{Final Aggregation} We assume that the new experts  $\{\mathcal{K}_1,\ldots,\mathcal{K}_P\}$ are conditionally independent given $y^*$ (see Figure \ref{fig.1.b}), which is not a strong assumption due to the process by which they were generated. The task is to find the distribution of new experts $\{\mathcal{K}_1,\ldots,\mathcal{K}_P\}$ and then find $p(y^*|\mathcal{D},x^*)$. The authors in \cite{Liu} showed that GRBCM provides consistent predictions under some mild assumptions, i.e. it can recover the true posterior distribution of $y^*$ when $n \to \infty$. Hence, we use the GRBCM aggregation method in each cluster by adding the global communication expert  $M_b$ to all clusters. For aggregating the new experts, we use either GPoE or GRBCM. Since the number of experts that are aggregated in each step is smaller than $M$, the computational cost of this scenario is smaller than the computational cost of GRBCM. Algorithm \ref{alg:1} depicts the aggregation process. 

\begin{algorithm}[H]
%\begin{algorithm*}
\caption{Aggregating Dependent Local Gaussian Experts}
\label{alg:1}
\begin{algorithmic}[1]
\REQUIRE {$\mu^*_M$, $\lambda$, P}
\STATE Calculate sample covariance S of experts' predictions 
\STATE Estimate $\hat{\Omega}$ using \eqref{eq:glasso}
\STATE Estimate $\mathcal{H}$ by performing spectral clustering \emph{SC}($\hat{\Omega}$, P)  
\STATE Obtain new experts $\{\mathcal{K}_1,\ldots,\mathcal{K}_P\}$ using GRBCM \eqref{eq:grbcm} 
\STATE Aggregate new experts using GPoE \eqref{eq:gpoe} or GRBCM \eqref{eq:grbcm}
\RETURN The estimated mean and variance of $p(y^*|\mathcal{D},x^*,{\mathcal{K}})$, i.e. $\mu^*_{\mathcal{K}}$ and $\Sigma^*_{\mathcal{K}}$
\end{algorithmic}
\end{algorithm}
%\end{algorithm*}

The following proposition gives our predictive distribution and its asymptotic properties.

\begin{Proposition}[\textbf{Predictive Distribution}]
Let $X$ be a compact, nonempty subset of $\mathcal{R}^{n \times D}$, $\mu^*_M=[\mu_1^*,\ldots,\mu_M^*]^T$ be the sub-models' predictions. We use $\{\mathcal{K}_1,\ldots,\mathcal{K}_P\}$ as defined in Algorithm \ref{alg:1}. We further assume that (i) $\lim_{n\to \infty} M = \infty$, (ii) $\lim_{n\to \infty} m_0 = \infty$, where $m_0$ is the partition size, and (iii) $\lim_{n\to \infty} |\mathcal{C}_i| = \infty,\; i=1,\ldots,P$, where $|\mathcal{C}_i|$ is the size of $i$'th cluster. The second condition implies that the original experts become more informative with increasing $n$, while the third condition means that the number of experts in each cluster increases. In addition, the third condition implies that $P \ll M$, which describes the dependency between the experts. \footnote{If we assume perfect diversity between experts (i.e., CI), then $P \approx M$. In this case, the consistency still holds due to the consistency of GRBCM but it is not a realistic assumption.}
Then the estimation based on Algorithm \ref{alg:1}, $y^*_{\mathcal{K}}$ is consistent, i.e.
\begin{equation} \label{eq:consistent_prediction}
  \begin{cases}
   \lim_{n\to \infty} \mu^*_{\mathcal{K}}=\mu^* \\
   \lim_{n\to \infty} \Sigma^*_{\mathcal{K}}=\Sigma^*.
  \end{cases}
\end{equation} 
\end{Proposition}

\textit{Proof:} The proof is straightforward due to the consistency of GRBCM. According to assumptions (ii) and (iii), when $n \to \infty$, each cluster returns a consistent predictor because the aggregation inside the clusters is based on GRBCM. Combining the consistent new experts $\{\mathcal{K}_1,\ldots,\mathcal{K}_P\}$ in Step 5 of Algorithm \ref{alg:1} leads to a consistent prediction. We provide here the proof for the variance, when GPoE is used in Step 5, and note that the proof for the mean is analogous. 

Let $\Sigma_{\mathcal{K}_i}^*$ be the covariance matrix of $\mathcal{K}_i$ which is obtained in Step 4 of Algorithm \ref{alg:1}, then the  aggregated precision of GPoE  (Step 5 of Algorithm \ref{alg:1}) is equal to  
\begin{align*}
\lim_{n\to \infty} (\Sigma_{\mathcal{K}}^*)^{-1} = & \lim_{n\to \infty} \sum_{i=1}^P \frac{1}{P} (\Sigma_{\mathcal{K}_i}^*)^{-1} = \sum_{i=1}^P \frac{1}{P} \lim_{n\to \infty}(\Sigma_{\mathcal{K}_i}^*)^{-1} \\ =& \sum_{i=1}^P \frac{1}{P} (\Sigma^*)^{-1} =(\Sigma^*)^{-1}, 
\end{align*}
where the first equality is based on the definition of GPoE with equal weights and the third one is due to the consistency of GRBCM.

In the next section, we showcase the importance of taking local experts' dependencies into account and the competitive performance of our approach using both artificial and real-world datasets.

\section{Experiments} \label{experiments}
The prediction quality of the proposed dependent Gaussian expert aggregation method (DGEA) is assessed in this Section. We showcase the importance of taking local experts' dependencies into account and the competitive performance of our approach using both artificial and real datasets. The quality of predictions is evaluated in two ways, standardized mean squared error (SMSE) and the mean standardized log loss (MSLL). The SMSE measures the accuracy of prediction mean, while the MSLL evaluates the quality of predictive distribution \cite{Rasmussen}. The standard squared exponential kernel with automatic relevance determination and a Gaussian likelihood is used. The experiments have been done in MATLAB using the GPML package\footnote{\url{http://www.gaussianprocess.org/gpml/code/matlab/doc/}}. The random partitioning method on the training dataset has been used in all experiments.

\subsection{Toy Example} \label{f_x}
The goal of our first experiment is to study the effect of dependency detection on the prediction quality and computation time. It is based on simulated data of a one-dimensional analytical function \cite{Liu},
\begin{equation}
f(x) = 5x^2\sin(12x) + (x^3 -0.5)\sin(3x-0.5)+4\cos(2x) + \epsilon,  \label{eq:toy_example}
\end{equation}
where $\epsilon \sim \mathcal{N}\left(0, (0.2)^2\right)$. We generated $n=10^4$ training points in $[0,1]$, and $n_t=0.1n$ test points in $[-0.2,-1.2]$. The data is normalized to zero mean and unit variance.

\begin{figure}[hbt!]
\centering
\subfloat[ SMSE]{\includegraphics[width=0.85\columnwidth]{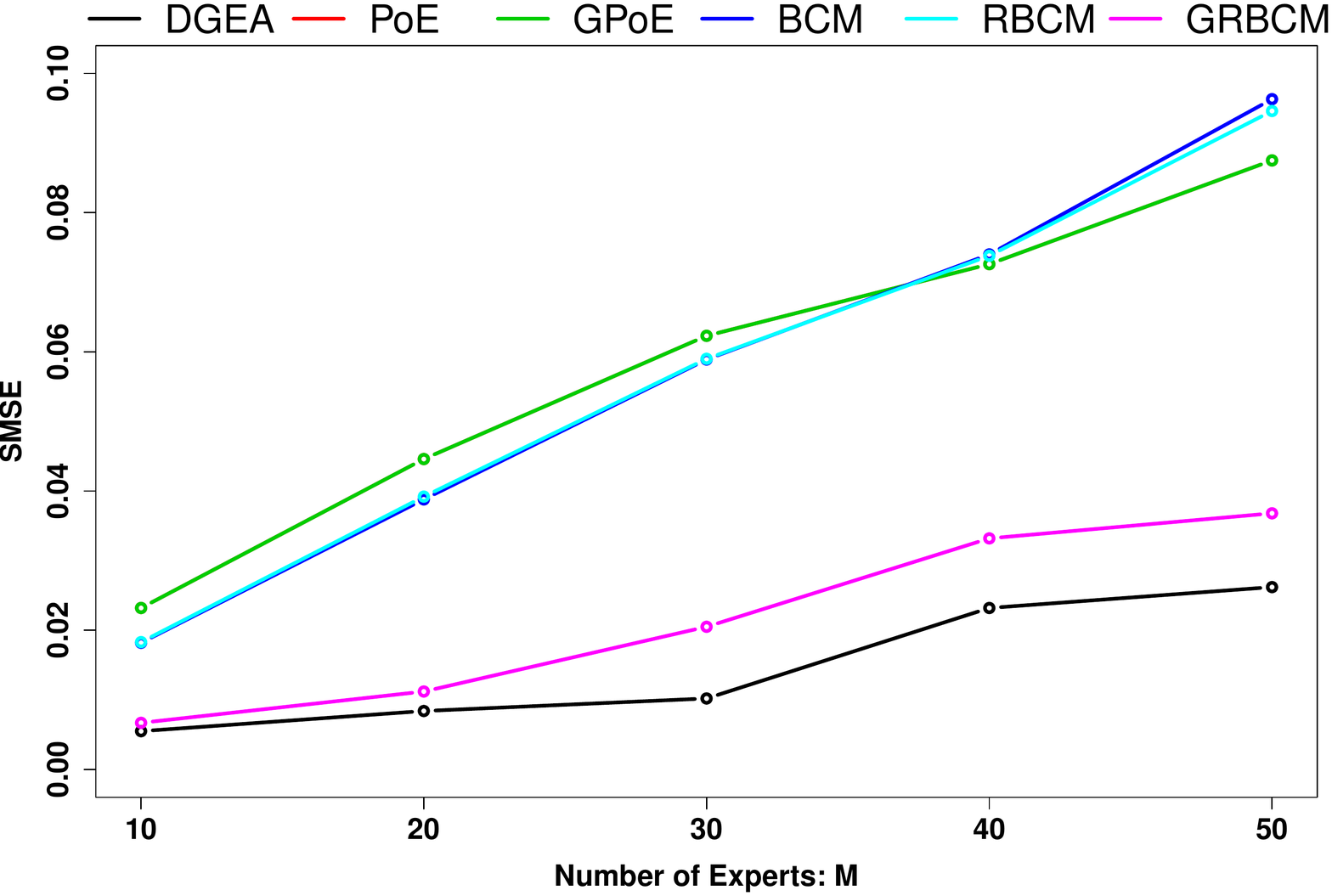}%
\label{fig.3.a}}
\hfil
\subfloat[MSLL]{\includegraphics[width=0.85\columnwidth]{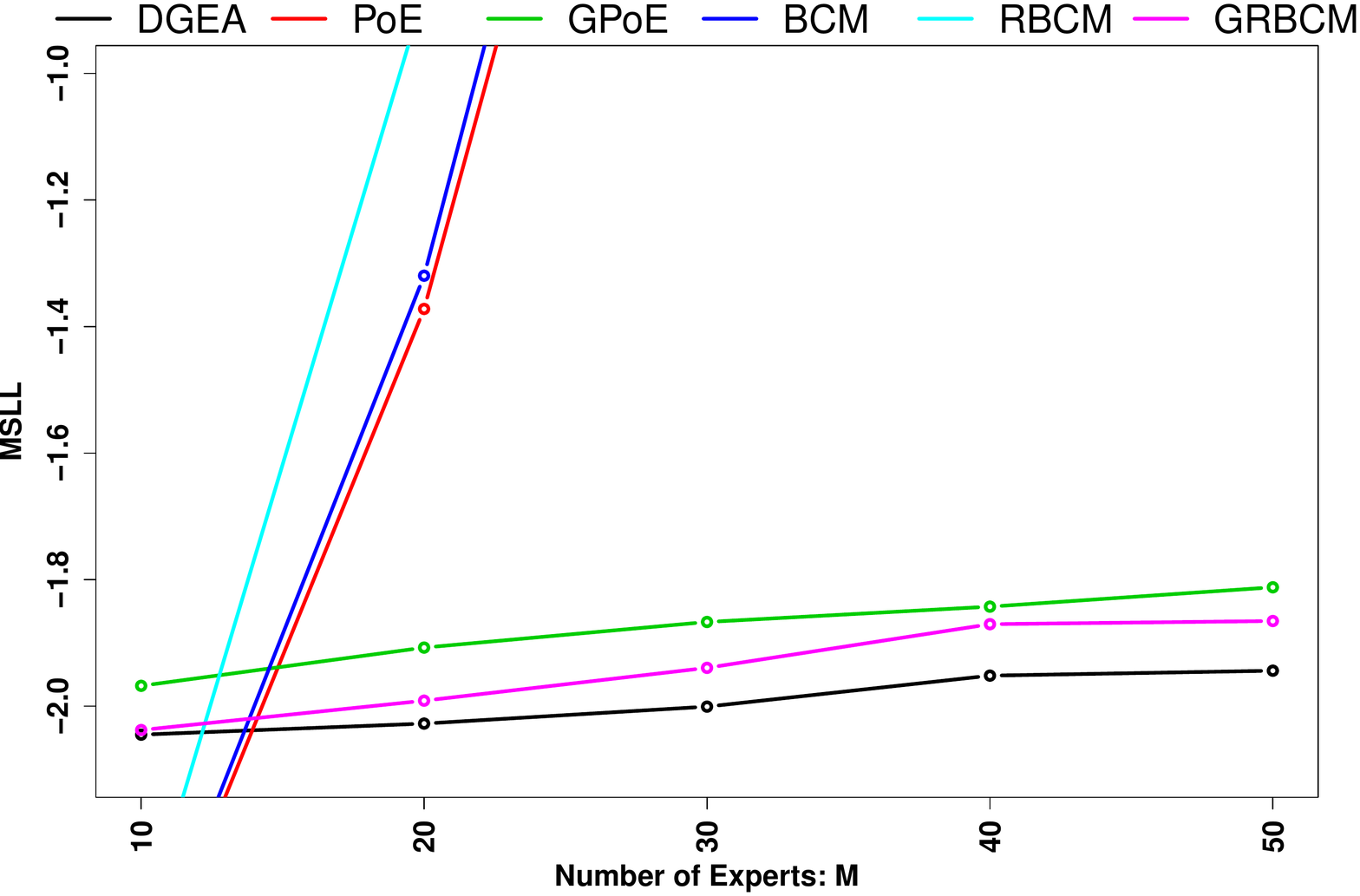}%
\label{fig.3.b}}
\hfil
\subfloat[Time]{\includegraphics[width=0.85\columnwidth]{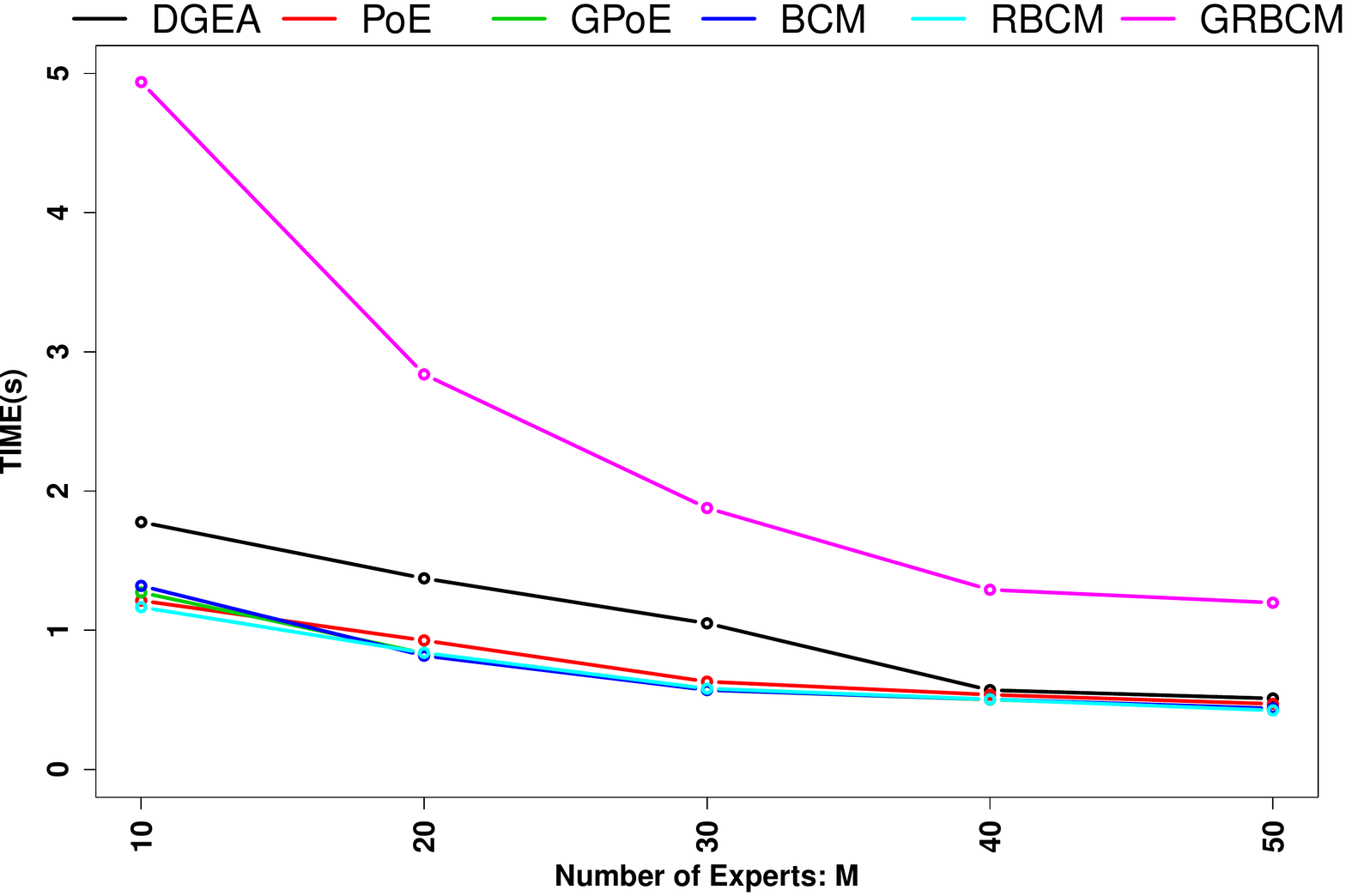}%
\label{fig.3.c}}
\caption{\textbf{Prediction quality} of different DGP methods with respect for different number of experts in the simulated data of the analytic function by \eqref{eq:toy_example}.}
\label{fig.3}
\end{figure}

We assigned 200, 250, 330, 500 and 1000 data points to each expert, which leads to 50, 40, 30, 20, and 10 experts respectively. Figure \ \ref{fig.3} shows the sensitivity of different DGP methods with respect to the change in the number of experts.

\begin{table*}[hbt!]
\caption{\textbf{Prediction quality} for various methods on \textit{Pumadyn}, \textit{Kin40k}, \textit{Sacros}, and \textit{Song} data. For both quality measure, i.e. SMSE and MSLL, smaller values are better.}
\label{table.1}
\vskip -5 mm
\begin{center}
\begin{small}
\begin{sc}
\begin{tabular}{lccccccccr}
\toprule
\multirow{2}{*}{} &
\multicolumn{2}{c}{\textit{Pumadyn}} &
\multicolumn{2}{c}{\textit{Kin40k}} &
\multicolumn{2}{c}{\textit{Sacros}} &
\multicolumn{2}{c}{\textit{Song}} & \\
\toprule
Model & SMSE  & MSLL  & SMSE & MSLL  & SMSE  & MSLL& SMSE  & MSLL  \\
\midrule
DGEA (Ours) & \textbf{0.0486} & \textbf{-1.5133} & \textbf{0.0538} & \textbf{-1.3025} &  \textbf{0.0269} & \textbf{-1.823} &  \textbf{0.8084}  & \textbf{-0.122}  \\
PoE    & 0.0505  & 4.8725 & 0.856& 2.4153 & 0.0311& 25.2807 & 0.8169 & 69.9464 \\
GPoE   & 0.0505 & -1.4936 & 0.0856& -1.2286  & 0.0311& -1.7756 & 0.8169 &\textbf{-0.123} \\
BCM    & 0.0499 & 4.6688 & 0.0818 & 1.6974 &  0.0308 & 24.868 & 10.4291& 44.1745 \\
RBCM   & 0.0498 & 12.1101 & 0.0772& 2.5256 & 0.0305& 61.5392 & 5.4373& 1.2089 \\
GRBCM  & 0.0511  & -1.488 &0.0544& -1.2785  & 0.0305& -1.4308 & 0.8268& 0.2073  \\
\bottomrule
\end{tabular}
\end{sc}
\end{small}
\end{center}
\vskip -0.1in
\end{table*}
The DGEA prediction is based on using GPoE in Step 5 of Algorithm \ref{alg:1}. Figure 
\ref{fig.3.a} depicts the SMSE values of the different SOTA methods. Since PoE and GPoE have the same SMSE value, the line of PoE is hidden in the plot. By increasing the number of experts, the prediction error rises, because we assign a smaller amount of observations to each expert, and therefore the quality of each expert decreases. While (G)PoE and (R)BCM return poor predictions, the DGEA and GRBCM present better results and DGEA has the smallest prediction error for the different numbers of experts.  

Figure \ref{fig.3.b} reveals the quality of the predictive distribution. The overconfident methods (PoE, BCM, and RBCM) return smaller MSLL for 10 experts. However, their MSLL value dramatically increases with growing $M$. As the authors in \cite{Bachoc} and \cite{Szabo} have shown, the GPoE is a conservative method and thus returns a higher quality. In Figure \ref{fig.3.b} we can see that its predictive distribution has a higher quality compared to PoE and (R)BCM. However, the DGEA represents an even higher quality for the predictive distribution for the different values of $M$. Figure \ref{fig.3.c} shows the computational costs of different methods. Comparing the running time of DGEA and GRBCM demonstrates that DGEA takes about half of the time of GRBCM, while its running time is almost indistinguishable from the running time of the most efficient methods, (G)PoE and (R)BCM.

\subsection{Realistic Datasets} \label{real_datasets}
In this section, we use four realistic datasets, \textit{Pumadyn}, \textit{Kin40k}, \textit{Sacros}, and \textit{Song}. The \textit{Pumadyn}\footnote{\url{https://www.cs.toronto.edu/~delve/data/pumadyn/desc.html}} is a generated 32D dataset with 7168 training points and 1024 test points. The 8D \textit{Kin40k} dataset \cite{Seeger} contains $10^4$ training points and $3\times10^4$ test points. The \textit{Sacros}\footnote{\url{http://www.gaussianprocess.org/gpml/data/}} is a 21D realistic medium-scale dataset with 44484 training and 4449 test points. 
The \textit{Song} dataset \footnote{\url{https://archive.ics.uci.edu/ml/datasets/yearpredictionmsd}} \cite{Bertin-Mahieux} is a 91D dataset with 515,345 instances which is divided into 463,715 training examples and 51,630 test examples. We extract the first $10^5$ songs from this dataset for training and keep the original set of 51,630 songs for testing. The random partitioning method has been used to divide the dataset into partitions and to generate the experts. The number of experts is 20 for \textit{Pumadyn} and \textit{Kin40k}, 72 for \textit{Sacros}, and 150 for \textit{Song}. For the \textit{Pumadyn} and \textit{Kin40k} datasets, 5 clusters, and for \textit{Sacros} and \textit{Song}, 10 clusters are used.  

Table \ref{table.1} depicts the prediction quality of different methods. In the \textit{Pumadyn}, \textit{Kin40k}, and \textit{Sacros} dataset, DGEA clearly outperforms the other methods. BCM and RBCM show lower prediction error compared to (G)PoE on these datasets, but their negative log-likelihood (MSLL) is quite large. Since GPoE provides conservative predictions and its posterior distribution converges to the true predictive distribution, it shows a nice performance with respect to the MSLL value, even better than the performance of GRBCM on the \textit{Pumadyn}, and \textit{Sacros} datasets. The drawback of PoE and (R)BCM methods can be seen in their MSLL values, which shows that their predictive distribution does not have competitive quality and tends to produce overconfident and inconsistent predictions, which has also been discussed by \cite{Szabo} and \cite{Liu}. With respect to the \textit{Song} dataset, DGEA and GPoE return better predictions. While GPoE performs a little bit better than DGEA with respect to  MSLL, DGEA has a lower prediction error. 

The GRBCM method returns different prediction qualities for different detests. Since in this work a new ensemble method is proposed for the non-parametric regression problem, the random partitioning is used, because in this case all Gaussian experts can cover the full sample space and work as global predictors. The quality of GRBCM is higher with respect to the disjoint partitioning, which is consistent with the results presented in~\cite{Liu}. But overall, the prediction quality of DGEA outperforms the other methods, which shows the importance of taking the experts' dependencies into account. 

\section{Conclusion}
In this work, we have proposed DGEA, a novel DGP approach which leverages the dependencies between experts to improve the prediction quality through local aggregation of experts' predictions. To combine correlated experts, comparable SOTA methods assume conditional independence between experts, which leads to poor prediction in practice. Our approach uses an undirected graphical model to detect strong dependencies between experts and defines clusters of interdependent experts. Theoretically, we showed that our new local approximation approach provides consistent results when $n \to \infty$. Through empirical analyses, we illustrated the superiority of DGEA over existing SOTA aggregation methods for scalable GPs. 

For future work, we identify two directions for further research. First, for the aggregated posterior, we integrated Gaussian graphical models into the generalized robust Bayesian committee machine and generalized product of experts. Another aggregation approach can be the latent variable graphical model, assuming that the final predictor is a latent variable within the graph that may improve the prediction quality by using interdependencies between experts while reducing time complexity. Second, the GGM relies on the assumption that all experts are jointly Gaussian and cannot be used to explain complex models with the non-Gaussian distribution. Therefore, finding a flexible and capable substitute for the GGM to capture the properties of GP experts is left to future work. 
\bibliographystyle{./IEEEtran}
\bibliography{./mybib}
\end{document}